\documentclass[11pt,a4paper]{article}
\usepackage[hyperref]{emnlp2018}
\usepackage{times}
\usepackage{latexsym}
\usepackage{graphicx}
\usepackage{tikz}
\usetikzlibrary{calc,trees,positioning,arrows,chains,shapes.geometric,%
  decorations.pathreplacing,decorations.pathmorphing,shapes,%
  matrix,shapes.symbols,fit,decorations,arrows.meta,%
  fit,shapes.misc}

\usepackage{url}

\aclfinalcopy 

\title{Reduce, Reuse, Recycle:\\ New uses for old QA resources.}

\author{Jeff Mitchell \and
  Sebastian Riedel \\
  University College London \\
  {\tt \{j.mitchell, s.riedel\}@cs.ucl.ac.uk} \\}

\date{}

\begin{document}
\maketitle
\begin{abstract}
We investigate applying repurposed generic QA data and models to a recently proposed relation extraction task. 
We find that training on SQuAD produces better zero-shot performance and more robust generalisation compared to the task specific training set. 
We also show that standard QA architectures (e.g. FastQA or BiDAF) can be applied to the slot filling queries without the need for model modification.
\end{abstract}

\section{Introduction}

Knowledge Base Population (KBP, e.g.: \citealp{riedeletal2013,sterckxetal2016}) attempts to identify facts within raw text 
and convert them into triples consisting of a subject, object and the relation between them. 
One common form of this task is slot filling \cite{surdeanuheng2014}, in which a knowledge base (KB) query, 
such as $place\_of\_birth(Obama,?)$ is applied to a set of documents and a set of slot fillers is returned. 
By converting such KB queries to natural language questions, 
\citet{levyetal2017} showed that a question answering (QA) system could be effectively applied to this task. 
However, their approach relied on a modified QA model architecture and a dedicated slot-filling training corpus.

Here, we investigate the utility of standard QA data and models for this task. 
Our results show that this approach is effective in the zero-shot and low-resource cases, 
and is more robust on a set of test instances that challenge the models' ability to identify 
relations between subject and object.

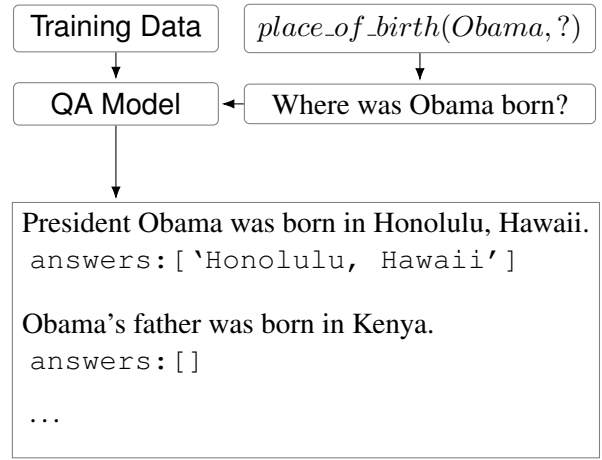
\begin{figure}[t]
\begin{center}
\begin{tikzpicture}[shorten >=1pt,draw=black!50, node distance=\layersep]

    \tikzstyle{box}=[rectangle, rounded corners=2]
    \node[draw, box, align=left, minimum width=0.35\columnwidth] (td) at (0,11) {\textsf{Training Data}};
    \node[draw, box, align=left, minimum width=0.35\columnwidth] (qa) at (0,10) {\textsf{QA Model}};
    \node[draw, box, align=left, minimum width=0.6\columnwidth] (sf) at (4,11) {$place\_of\_birth(Obama,?)$};
    \node[draw, box, align=left, minimum width=0.6\columnwidth] (sfq) at (4,10) {Where was Obama born?};
    
    \path[draw=black,solid,-{Latex}] (td) edge (qa);
    \path[draw=black,solid,-{Latex}] (sf) edge (sfq);
    \path[draw=black,solid,-{Latex}] (sfq) edge (qa);
    
    \node[draw, align=left] (out) at (2.5,7) {President Obama was born in Honolulu, Hawaii.\\
                                            ~\texttt{answers:[`Honolulu, Hawaii']}
                                            \vspace{2ex}\\
                                            Obama's father was born in Kenya.\\
                                            ~\texttt{answers:[]}
                                            \vspace{1ex}\\
                                            ~\ldots \vspace{2ex}};

    \draw[black,-{Latex}] (qa.south) to (qa.south|-out.north);

\end{tikzpicture}
\caption{QA for KBP overview.}\label{introfigqakbp}
\end{center}
\end{figure}

Figure \ref{introfigqakbp} gives an overview of using QA on the slot-filling task. 
Starting at the top right, a KB query is translated into a natural language question, 
which can then be fed into a QA model that has been trained on an appropriate resource. 
When applied to a set of texts, this model needs to predict the correct answer within each text, 
including the possibility that a text contains no answer. 
Within this framework, we consider different models and training and test datasets, 
but we keep the translation of KB queries into natural language questions fixed, 
based on the crowd-sourced templates used by \citet{levyetal2017}.

\section{Performance on the original task}\label{exp1}

In our first experiment, we examine the utility of a standard QA dataset 
as training data for the slot-filling model of \citet{levyetal2017}. 
Their zero-shot model generalised from seen relations to unseen relations 
by translating all relations into natural language question templates, such as \emph{Where was XXX born?} for the relation $place\_of\_birth$. 
Identifying an instance of such a relation in text is then equivalent to finding an answer to the relevant question template, 
instantiated with the appropriate entity. 
However, such a model also needs to be able to identify when no answer is found in the text, 
and to achieve this they trained a slightly modified version of BiDAF \cite{seoetal2016} 
on both positive examples, containing answers, and negative examples, without answers. 
These examples were derived from a pre-existing relation extraction resource, 
as their intention was to show the utility of the QA model.

In this section, we evaluate whether the same model trained on QA data, specifically SQuAD \cite{rajpurkaretal2016}, can be applied to the relation extraction task. 
We first investigate the zero-shot case, where no examples of the relations are available, 
and then evaluate how performance improves as more data becomes available.

\begin{table}[t]
\centering
\begin{tabular}{l|ccc}
Training Data & P    & R    & F1 \\
\hline
UWRE          & 0.43 & 0.36 & 0.39 \\
SQuAD         & 0.44 & 0.41 & 0.43 \\
\end{tabular}\caption{Zero-Shot Precision, Recall and F1 on the UWRE relation split test set.}\label{exp1tabrels}
\end{table}

\paragraph{Data}

We compare two sources of training data: 
The University of Washington relation extraction (UWRE) dataset created by \citet{levyetal2017}
and
the Stanford Question Answering Dataset (SQuAD) created by \citet{rajpurkaretal2016}.

The UWRE data is derived from WikiReading \cite{hewlettetal2016}, which is itself derived from WikiData \cite{vrandecicetal2012},
and consists of a set of positive and negative examples for relation extraction from Wikipedia sentences. 
Each instance consists of an entity, a relation, a question template for the relation 
and a sentence drawn from the wikipedia article for that entity which may or may not answer the question. 
Under the assumption that each relation triple found in a Wikipedia info-box is also expressed in the text of its article, 
the positive examples contain the first sentence from the article that contains both the subject and object of the triple. 
The negative examples also contain the subject entity of the relation, but express a different relation.

\begin{figure}[t]
\centering
\includegraphics[scale=0.4]{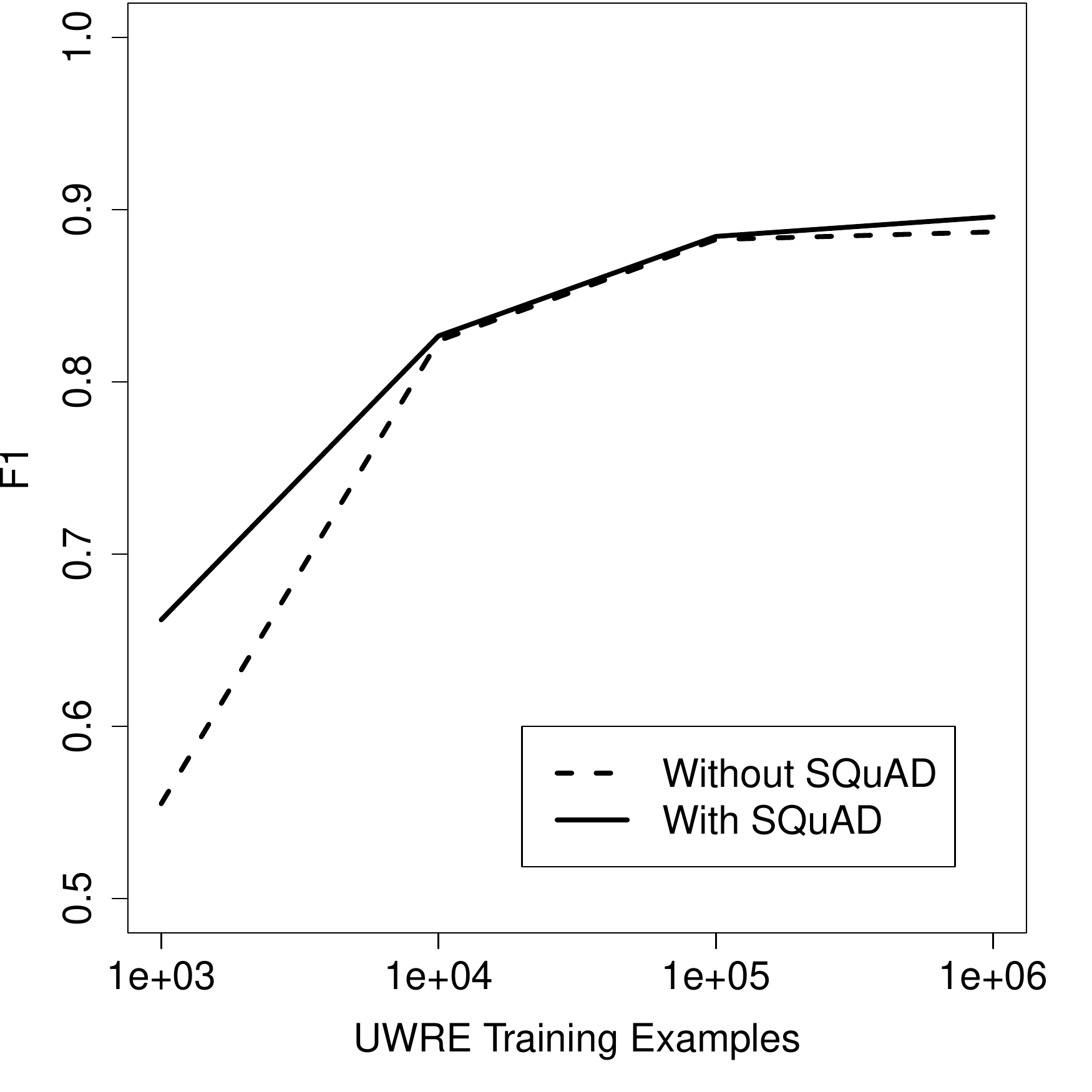}
\caption{F1 on the UWRE entity split test set.}\label{exp1figents}
\end{figure}

\citet{levyetal2017} provide a number of train/dev/test splits, to allow them to evaluate a variety of modes of generalisation. 
Here we use the relation and entity splits. 
The former tests the ability to generalise from one set of relations to another, 
i.e. to do zero-shot learning for the unseen relations in the test set.
In contrast, the latter tests on the easier task of generalising from one set of entities to another for the same set of relations. 
We use this dataset to investigate how having access to various quantities of data about the test set relations changes performance.

To build a dataset using SQuAD \cite{rajpurkaretal2016}, 
we construct negative examples by removing sentences that contain the answer, based on the spans provided by the annotators. 
In other words, we are left with the original question and a paragraph relevant to the topic of that question, 
but which typically no longer contains sentences answering it. 
Alongside these negative examples, we also retain the original SQuAD instances as positive examples. 
This process is applied to both the train and dev sets, 
allowing us to evaluate a model that uses only question answering data at training time.

We also construct a series of datasets that combine increasing quantities of the UWRE entity split training set into the SQuAD training set,
to evaluate the benefits of SQuAD when dedicated relation extraction data is limited. 
Random samples of $10^3$, $10^4$, $10^5$ and $10^6$ UWRE instances are added to our SQuAD training set, 
while leaving the SQuAD dev dataset untouched.

\paragraph{Models}

We employ the same modified BiDAF \cite{seoetal2016} model as \citet{levyetal2017}, 
which uses an additional bias term to allow the model to signal when no answer is predicted within the text.

\paragraph{Evaluation}

Following the approach of \citet{levyetal2017}, we report F1 scores on the answers returned by the model. 
Under this measure, predicting correctly that a negative instance has no answer 
does not contribute to either precision or recall. 
However, returning an answer for such an instance does reduce precision.

\begin{table}[t]
\centering
\begin{tabular}{l|r}
Training Data & Acc \\
\hline
UWRE          & 0.02 \\
UWRE+         & 0.73 \\
SQuAD         & 0.83 \\
\end{tabular}\caption{Accuracy on the challenge test set of models trained on SQuAD and the UWRE and UWRE+ entity split data.}\label{exp2tabents}
\end{table}

\paragraph{Results}

Table \ref{exp1tabrels} reports the F1 scores for zero-shot relation extraction on the relation split test set, 
using models trained on the original UWRE and SQuAD datasets. 
As can be seen,
BIDAF is actually more effective at answering the questions for the unseen relation types in the UWRE test set
when it is trained on a standard QA dataset, rather than a dedicated relation extraction dataset.

Figure \ref{exp1figents} plots how performance improves as more data becomes available about the relations in the entity split test set. 
We compare training purely on UWRE instances to those same instances combined with the whole SQuAD dataset.
As can be seen, when only small amounts of relation extraction data is available, 
combining this with the QA data gives a substantial boost to performance.

\paragraph{Discussion}

The SQuAD trained model appears to be effective in the limited data and zero-shot cases, 
but contributes little when large numbers of examples of the relations of interest are available. 
In this case, the dedicated relation extraction model is able to achieve an F1 of around 90\%, with or without augmentation with SQuAD. 
This level of performance suggests that such a model would be accurate enough for practical applications.
However, test set performance may not be a reliable indicator of the model's ability 
to generalise to more challenging examples \cite{jialiang2017}.

\section{Generalisation to a challenge test set}\label{exp2}

\begin{figure}
\centering
\includegraphics[scale=0.4]{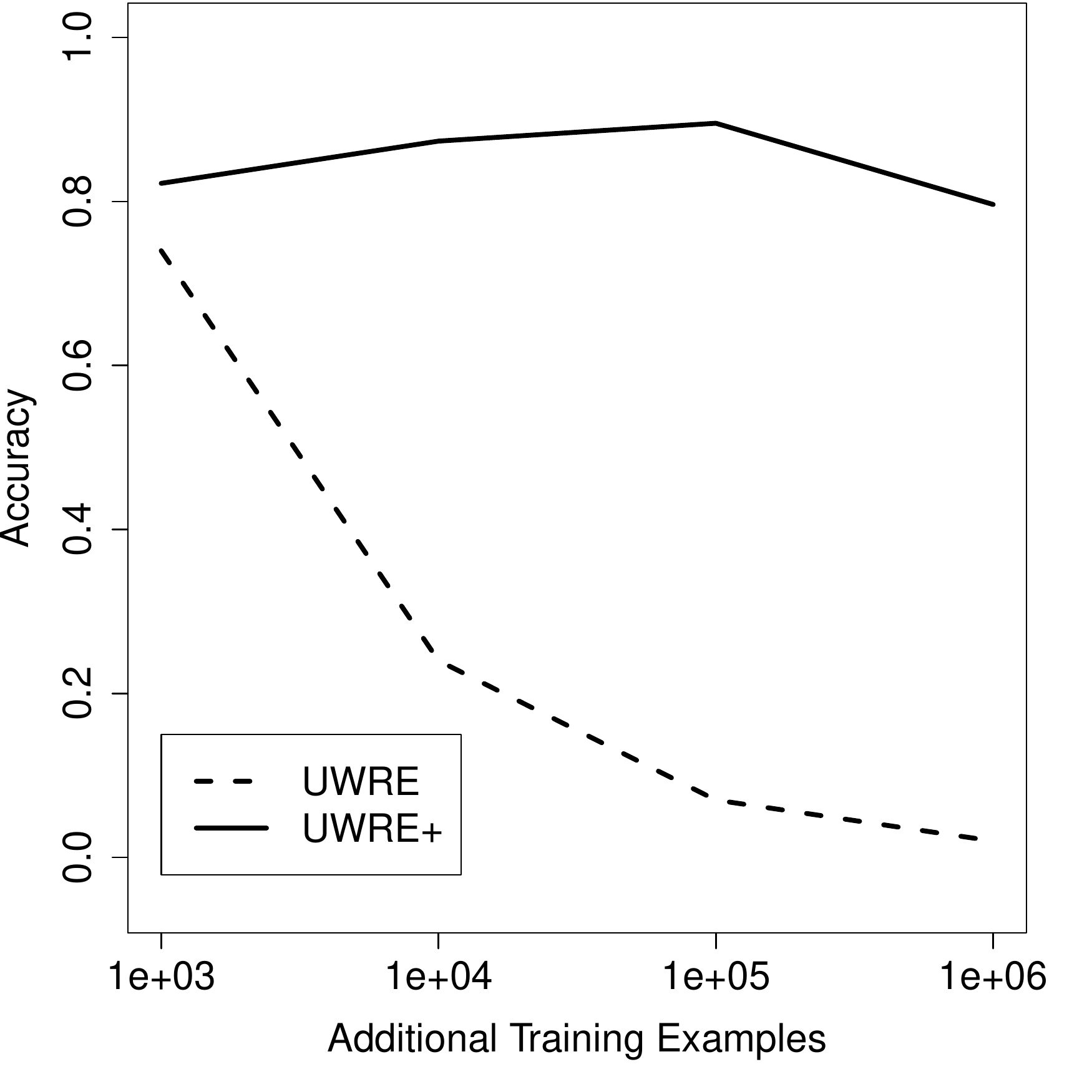}
\caption{Accuracy on the challenge test set as SQuAD is augmented with increasing amounts of entity split training data.}\label{exp2figents}
\end{figure}

In this second experiment, we want to test the ability of the models decribed above to generalise 
to data beyond the UWRE test set. 
In particular, we want to verify that the BiDAF model is able to recognise the assertion of a relation between 
the entity and the answer, rather than just recognising an answer phrase of the right type.

\paragraph{Data}

We construct a challenge test set of negative examples based on sentences which are about the wrong entity 
but which do contain potential answers that are valid for the question and relation type. 
Thus, each positive example from the original UWRE entity split test set is turned into a negative example by pairing 
the sentence with an equivalent question about another entity. 
A model that has merely learnt to identify answer spans of the right form, 
irrespective of their relation to the rest of the sentence, 
is likely to return the original span rather than recognise that the sentence no longer contains an answer.

We then build new train, dev and test sets (UWRE+) from the original entity split datasets 
in which half the original negative instances have been replaced with these more challenging instances. 
As before, a series of datasets combining SQuAD with increasing amounts of this new data is also constructed.

\paragraph{Models}

We re-use the UWRE and SQuAD trained models in addition to training on the UWRE+ datasets described in the previous section.

\paragraph{Evaluation}

Here, F1 is not an appropriate measure, as there are no positive instances in the challenge data. 
Instead, we use accuracy of the predictions, which in this case is just the number of `no answer' predictions 
divided by the total number of instances.

\paragraph{Results}

Table \ref{exp2tabents} reports the accuracy of predictions on the challenge test set of negative examples. 
Although the original UWRE model achieved an F1 of around 90\% on the unmodified entity split test set, 
here it only manages to get 2\% of its predictions correct. 
In contrast, the modified UWRE+ training data results in a model that is much more accurate, predicting over 70\% of the negative examples correctly.
Nonetheless, the performance of the SQuAD trained model is stronger still, even without modification to address this problem.

Figure \ref{exp2figents} shows the accuracy on the challenge test set as increasing quantities of relation extraction instances are added to SQuAD. 
Looking first at the effect of adding the original UWRE training instances, 
performance drops dramatically as the size of this expansion increases. 
In contrast, as the quantity of UWRE+ data grows, performance improves, peaking at around 100,000 instances, 
which is around the same size as SQuAD.

\paragraph{Discussion}

The results on our challenge test set suggest that the model does not learn to examine the relation between the answer span and the relation subject unless the training data requires it. 
In the case of SQuAD, the multi-sentence paragraph structure around the answer  provides enough potential distractors to overcome this issue.

Other models may show different patterns of strength and weakness, 
but to be able to investigate and exploit further QA systems quickly 
would require a means of producing `no answer' predictions without the need to modify the model implementation.

\section{Using an unmodified QA model for slot filling}

\begin{table}[t]
\centering
\begin{tabular}{l|ccc}
Model         & P    & R    & F1 \\
\hline
BiDAF         & 0.40 & 0.34 & 0.37 \\
FastQA        & 0.49 & 0.19 & 0.28 \\
\end{tabular}\caption{Zero-shot Precision, Recall and F1 on the UWRE relation split test set.}\label{exp3tabrels}
\end{table}

\citet{levyetal2017} modify the BiDAF architecture to produce an additional output 
representing the probability that no answer is present in the text. 
In this experiment, we investigate whether it is possible to adapt a QA model to the slot filling task 
without having to understand and modify its internal structure and implementation. 
Our approach merely requires prefixing all texts with a dummy token 
that stands in for the answer when no real answer is present.

\paragraph{Data} 
We train our models on a modified version of SQuAD, which has been augmented with negative examples by removing answer spans, 
as described in Section \ref{exp1}, and then had the token \texttt{NoAnswerFound} 
inserted into every text and as the answer for the negative examples, as described above.

\paragraph{Models}
We train both BiDAF \cite{seoetal2016} and FastQA \cite{weissenbornetal2017} models on the modified SQuAD training data, using their standard architectures and hyperparameters.

\paragraph{Evaluation}
We evaluate F1 on the same zero-shot evaluation considered in Section \ref{exp1} 
and also accuracy on the challenge test set from Section \ref{exp2}.

\paragraph{Results}

Table \ref{exp3tabrels} reveals that the unmodified BiDAF model is almost as effective as 
the \citet{levyetal2017} model in terms of zero-shot F1 on the original UWRE test set. 
In contrast, FastQA's performance is substantially worse. 

However, Table \ref{exp3tabents} reveals that FastQA is extremely accurate on the challenge test set, 
while BiDAF's performance is comparable to the modified model trained on SQuAD.

\paragraph{Discussion} The unmodified BiDAF and FastQA architectures have complementary strengths on the two evaluations. 
FastQA's strong performance on the challenge instances may be related to its use of binary features 
indicating whether a word was present in the question.

\begin{table}[t]
\centering
\begin{tabular}{l|r}
Model & Acc \\
\hline
BiDAF         & 0.82 \\
FastQA        & 0.99 \\
\end{tabular}\caption{Accuracy on the challenge test set.}\label{exp3tabents}
\end{table}

\section{Conclusion}

We showed that standard QA models and data can be easily reused on the slot-filling task, 
using some straightforward data pre-processing. 
These recycled models were reasonably effective in the reduced data regime 
and robust on a new test set containing challenging examples.


\section*{Acknowledgments}

This work has been supported by the European Union H2020 project SUMMA (grant No. 688139), and by an Allen Distinguished Investigator Award.

\bibliography{wiki}
\bibliographystyle{acl_natbib}

\end{document}